\documentclass{article}

\usepackage{arxiv}
\usepackage[section]{placeins}

\usepackage[utf8]{inputenc} 
\usepackage[T1]{fontenc}    
\usepackage{hyperref}
\hypersetup{colorlinks,allcolors=black}
\usepackage{url}            
\usepackage{booktabs}       
\usepackage{amsfonts}       
\usepackage{nicefrac}       
\usepackage{microtype}      
\usepackage{lipsum}		
\usepackage{graphicx}
\usepackage{natbib}
\usepackage{doi}

\title{Exploratory Data Analysis of Urdu Poetry}

\author{ \hspace{1mm}{Shahid Rabbani}\thanks{Email:shahid.rabbani@ku.ac.ae} \\
	Mechanical Engineering Department\\
	Khalifa University\\
	Abu Dhabi, UAE \\
	\And Zahid Ahmed Qureshi
	 \\
	Department of Mechanical Engineering\\
	UAE University\\
	Alain, UAE \\
}

\renewcommand{\shorttitle}{\textit{arXiv} }

\hypersetup{
pdftitle={Exploratory Data Analysis of Urdu Poetry},
pdfsubject={q-bio.NC, q-bio.QM},
pdfauthor={Shahid Rabbani},
pdfkeywords={First keyword, Second keyword, More},
}
\begin{document}
\maketitle

\begin{abstract}
The study presented here provides numerical insight into \emph{ghazal}- the most appreciated genre in Urdu poetry. Using  48,761 poetic works from 4,754 poets produced over a period of 800 years, this study explores the main features of Urdu \emph{ghazal} that make it popular and admired more than other forms. A detailed explanation is provided as to the types of words used for expressing love, nature, birds, and flowers etc. Also considered is the way in which the poets addressed their loved ones in their poetry. The style of poetry is numerically analyzed using Multi Dimensional Scaling to reveal the lexical diversity and similarities/differences between the different poetic works that have drawn the attention of critics, such as Iqbal and Ghalib, Mir Taqi Mir and Mir Dard. The analysis produced here is particularly helpful for research in computational stylistics, neurocognitive poetics, and sentiment analysis. 
\end{abstract}


\section{Introduction}
Urdu language has more than 230 million speakers across the globe. It is ranked 10th with respect to the number of speakers among all the living languages (\cite{Top200}). Considering that there are presently approximately 7139 languages spoken in the world, this is an impressive ranking.The majority of its speakers come from South Asian countries such as Pakistan, India, Bangladesh, etc. Since the diaspora of these South Asian countries is also distributed in large numbers across the globe, this makes Urdu a recognizable language in several other countries with the most famous ones being the Gulf countries like UAE, Saudi Arabia, Oman, and Qatar, to name a few. In fact, a new lingua franca, Urdubic, is emerging in the Gulf countries whose linguistic composition is defined by the reduced and simplified forms of Arabic and Urdu (\cite{Hussain2020}]).

	From a historical perspective, Urdu has been pre-dominantly derived from an amalgamation of several languages with the most famous being Persian, Arabic, Sanskrit and Turkish languages. During the days of Mughal Dynasty’s rule in India, Urdu rose to its prominence. A number of poets also made their mark and became iconic names in Urdu poetry. The Two School of Urdu Poetry Theory, perhaps the most prevalent in Urdu literary criticism, holds that the Delhi School and the Lucknow School comprise the bulk of classical poetry. Two famous school of Urdu literature came into existence shortly afterwards i.e., Delhi and Lucknow, the then great centers of Urdu culture. \emph{Dehlavi} poetry (the poetry written by poets belonged to Delhi), considered by critics to be truer to the Persian literary tradition than the poetry of Lucknow, and is described as emphasizing mystical concerns, Persian styles of composition, and a straightforward, melancholy poetic diction. \emph{Lakhnavi} poetry (emerged from Lucknow) by contrast, is characterized as sensual, frivolous, abstruse, flashy, even decadent (\cite{Petievich1986}).  It is also pertinent to highlight that the prominence of Urdu poets did not remain limited to South Asia. In fact, a number of poets (for example, Muhammad Iqbal, Ghalib and Faiz) gained what can be termed as international recognition. Their poetic works have been translated into many languages and is being taught in various international universities outside South Asia as a part of their curriculum in language departments.

	The advent of computational tools has emerged as a robust tool to process and analyze huge data sets with efficient algorithms. In particular, Exploratory Data Analysis (EDA) has also seen rapid developments. EDA refers to the critical process of performing deep investigations on data so as to discover patterns, to spot anomalies, to test hypothesis and to check assumptions with the help of summary statistics and graphical representations (\cite{IBM}). Once EDA is performed and insights are drawn, its features can then be used for more sophisticated data analysis or modeling, including machine learning. The exploratory analysis of any language acts as dissection tool to understand its key features. However, it is pertinent to highlight that classical exploratory analysis relied upon rudimentary techniques like scavenging a complete set of work of one poet to comment upon his style of writing.  In order to portray an anatomy of several poets and all their published works, colossal amounts of time and energy would be required

Through the development of sophisticated data processing and analysis tools, data is discovering its new value with unprecedented applications. Specially, with the advent of Artificial Intelligence (AI) and Machine Learning (ML), the anatomy of a language and the understanding of literary characteristics of huge data sets (for example, several hundred poets and several thousand poems) became much easier. Unfortunately, South Asian languages and in particular Urdu though, have received very less attention in this regard. Anwar et al. (\cite{Anwar2006}) provides a classical survey of Urdu language processing while Daud et al. (\cite{Daud2017}) provides a contemporary survey of Urdu language processing. Shaikh et al. (\cite{Shaikh}) explained the working of system that translated text from English to Urdu and vice-versa with AI. It used ISO/IEC 10646 standard (\cite{ITset}) Urdu Unicode with open-type fonts and character controls in Operating System. Authorship attribution in Urdu language using AI techniques has also been studied; albeit a bit sparsely. Anwar et al. (\cite{Anwar2018}) performed an empirical study on forensic analysis of Urdu text using Latent Dirichlet Allocation (LDA) based authorship attribution. They presented a unified approach for intelligent association analysis of text of how much a piece of text is related to a person with respect to his stylometric writing features. Ashraf et al. (\cite{Ashraf2010}) presented an approach to develop an Automatic speech recognition system for Urdu language. The proposed system utilized statistical based Hidden Markov Model for development  for a relatively small sized vocabulary, i.e., 52 isolated most spoken Urdu words. Singh (\cite{Singh2012}) devised a named entity recognition system for Urdu. Sentiment Analysis of widely spoken languages such as English and Chinese have remained the focus of researchers’ attention but resource poor languages such as Urdu have been mostly ignored (\cite{Mukhtar2018}). Mukhtar and Khan (\cite{Mukhtar2018}) devised an algorithm for Urdu sentiment analysis using supervised machine learning approach. They acquired data sets from various blogs of fourteen different genres which were later annotated with the help of human annotators. Three well-known classifiers, i.e., Support Vector Machine , Decision tree and k-Nearest Neighbor (kNN) were tested, their outputs were compared and finally their results were improved in several iterations after taking a number of steps that included stop words removal, feature extraction, identification and extraction of important features. Khan et al. (\cite{Khan2021}) furthered the previous study (\cite{Mukhtar2018}) by conducting a Urdu sentiment analysis with deep learning methods. Majeed et al. (\cite{Majeed}) developed an algorithm for emotion detection in roman Urdu text using machine learning. In particular to Urdu poetry, Dar (\cite{Dar}) performed a study on author attribution in Urdu poetry. In her study, poetic collection (a total of 11406 couplets) from the works of three well-known Urdu poets (Muhammad Iqbal, Mirza Ghalib and Faiz Ahmed Faiz) were collected from reputable non-social Urdu poetry websites notably Rekhta (\cite{Rekhta}) and UrduLibrary (\cite{Urdulib}). After collection of the dataset, pre-processing was done by cleaning of the dataset i.e., removing commas, colon, semicolon, etc. After pre-processing, various classifiers (Multinomial Naive Bayes, Multilayer Perceptron, MLP pre-trained model word2vec, Support Vector Classifier RBF), were used to perform training and testing on the data set. The best results included Support Vector Classifier with the best F1-score at 82.67\% and the highest accuracy of 82.85/\%. Similarly, Rao and Ahmed (\cite{Rao}) presented a machine learning system to find the optimal configuration for poet attribution for Urdu for couplets. They argued that the task was more difficult than the general task of author attribution, as the number of words in verses and poems are usually less than the number of articles present in author attribution datasets. They applied classification algorithms with different sets of feature configurations to run several experiments and found that the system performed best when support vector machine using a combination of unigram and bigram were used. Their best reported system returned an accuracy of 88.7\%.

	With the above literature survey in consideration, it can be readily argued that there is a clear absence of a detailed and comprehensive EDA of Urdu poetry. In this work, we performed an extensive exploratory analysis of Urdu language. A total of 48,761 poetic works of 4,754 poets were analyzed in total that represents a huge sample from the entire existing well-established Urdu poetry.

\section{Method}
\label{sec:method}
Data is collected form the popular Urdu poetry platform \emph{Rekhta} [\cite{Rekhta}] which is considered to be the largest collection of Urdu poetry compiled in digital fonts. This platform consists of Urdu poetry of different genre including \emph{ghazal, nazm, shair and rubai} written in Urdu, Hindi and Roman transliteration. Without any preconceived bias,  all the \emph{ghazals} written in Urdu font were collected from the website and were stored  to one text file. Details of the collected data is given in Table \ref{tab:data_summary}~Cleaning operation was performed to all remove non-urdu and non-alphabetic characters which was followed by text normalization. Once data was collected, cleaned and normalized,  the corpus was analyzed for different features presented in below section.

\begin{table}[htbp]
\caption{Summary of Data}
\begin{center}
\begin{tabular}{ll}
\hline
Feature                      & Count   \\ \hline
Total number of Ghazals      & 48,761   \\
Total Number of poets        & 4,754    \\
Total number of verses (shair)                & 687,993  \\
Total number of Words        & 5,599,002 \\
Total number of unique words & 33,676   \\ \hline
\end{tabular}
\end{center}
\label{tab:data_summary}
\end{table}

\section{Results}
\label{sec:results}
 
\subsection{Bigrams and Trigrams}

\begin{figure}[htbp]
\begin{center}
\includegraphics[scale=0.42]{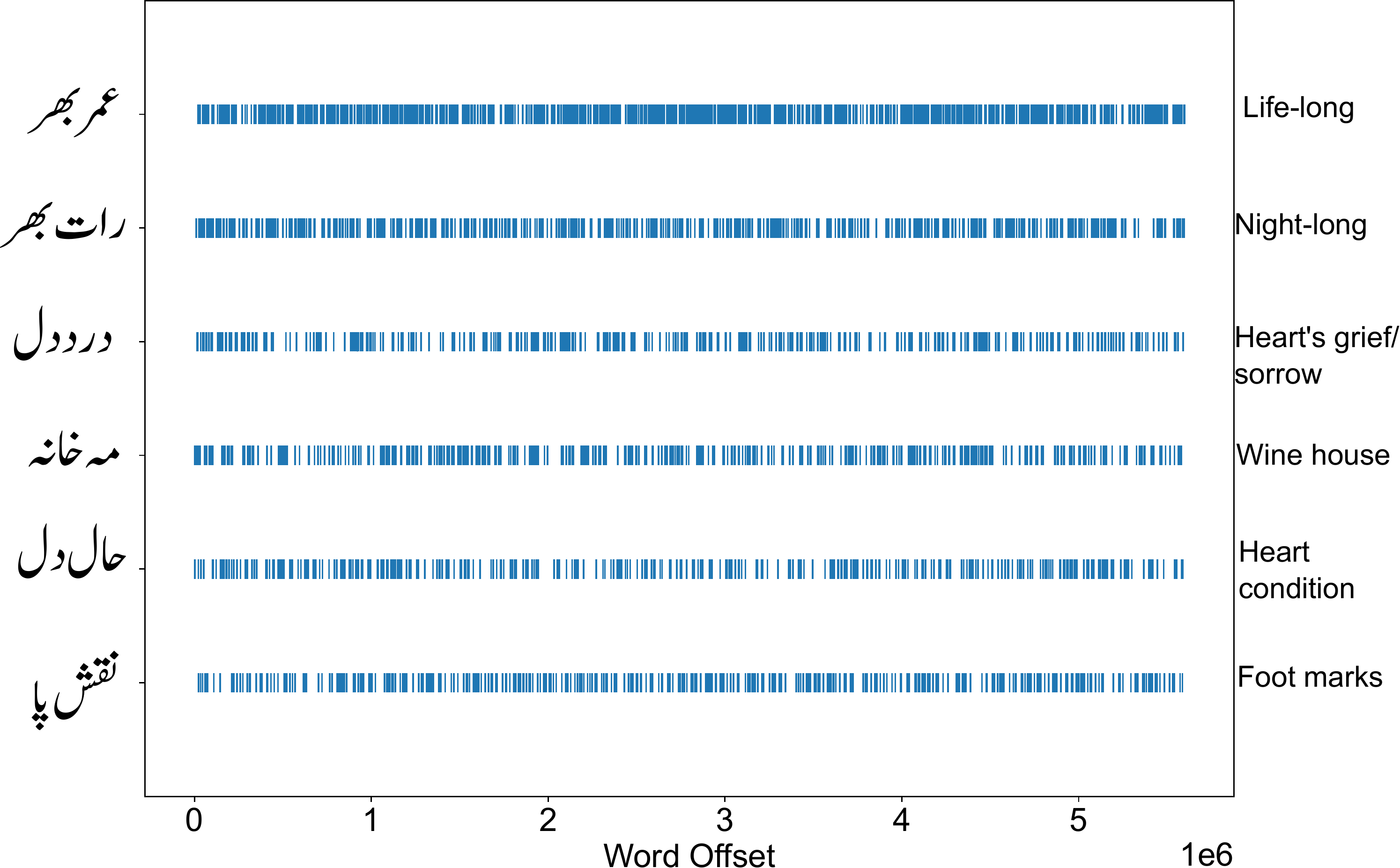}\\
\end{center}
\caption{Lexical dispersion plot of most frequent bigrams in order from top to bottom.}
\label{fig:lex_bigrams}
\end{figure}

\begin{figure}[htbp]
\begin{center}
\hbox{\hspace{10ex}\includegraphics[scale=0.42]{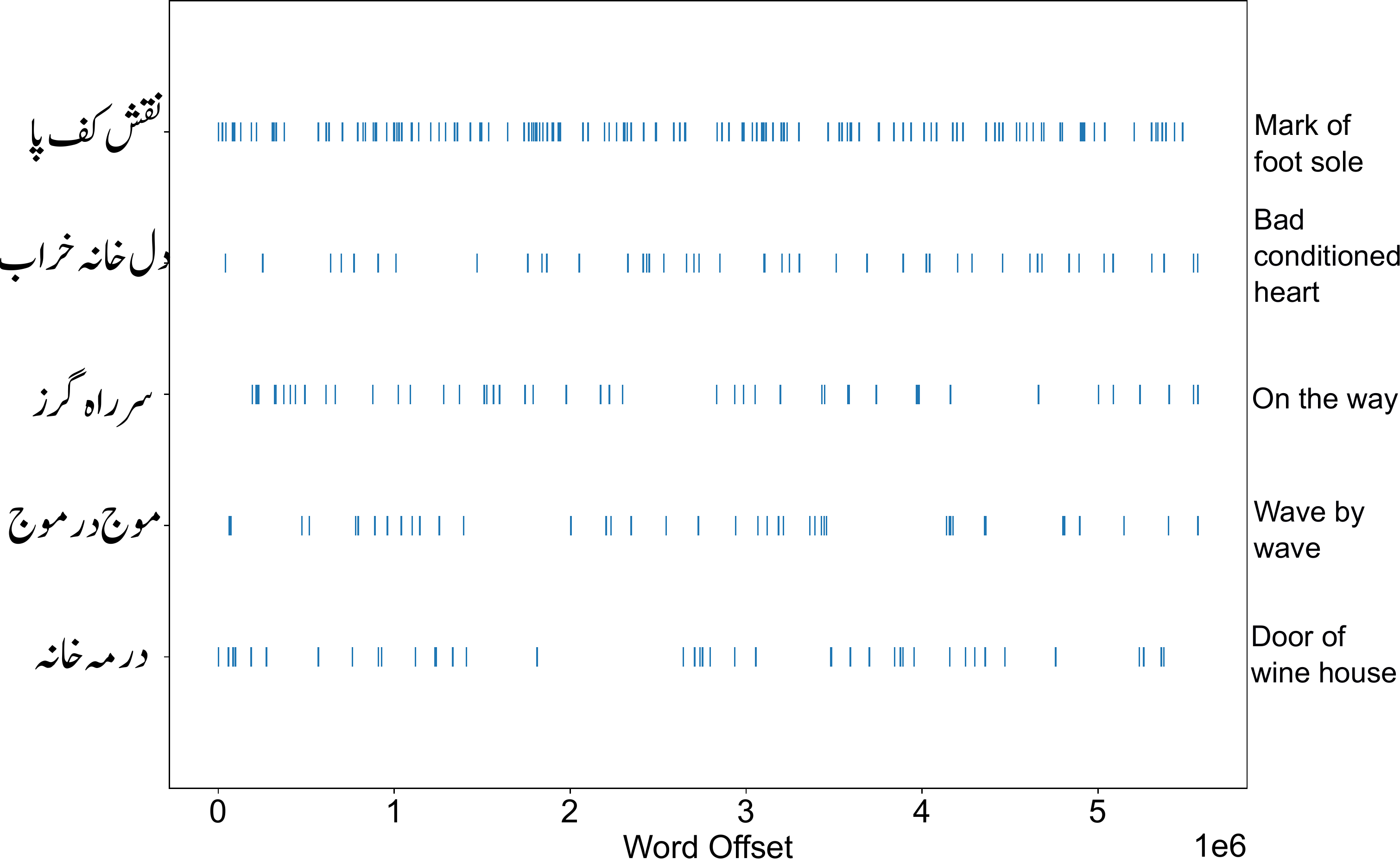}}
\end{center}
\caption{Lexical dispersion plot of most frequent trigrams in order from top to bottom.}
\label{fig:lex_trigrams}
\end{figure}

The findings of the exploratory analysis are presented in this section.
Figure \ref{fig:lex_bigrams} and Figure \ref{fig:lex_trigrams} depict the lexical dispersion plots of the most frequently appearing bigrams and trigrams (in order from top to bottom).

In the case of bigrams (Figure \ref{fig:lex_bigrams}), it can be readily seen that the elements of romanticism and longing for love have inundated the Urdu poetic works under consideration. The bigram like \emph{umar bhar} (life-long), \emph{raat bhar} (night-long), \emph{dard-e-dil} (heart’s grief/sorrow), \emph{ma'y-khana} (wine house), \emph{naqsh-e-pa} (foot marks) etc. appeared very frequently in the poetry. As these bigrams are often associated with nostalgia, romanticism and longing for lost-love and the frequent usage of these bigrams becomes rather obvious. Same is the case with the trigrams (Figure \ref{fig:lex_trigrams}) where there is a clear prevalence of such words expressed with more sophistications; such as \emph{naqsh-e-kaf-e-pa} (mark of foot sole), \emph{dil-e-khana kharab} (bad-conditioned heard) and \emph{dar-e-ma'y khana} (Door of wine house).

\subsection{Analysis of words usage}

\begin{figure}[htbp]
\begin{center}
\includegraphics[scale=0.42]{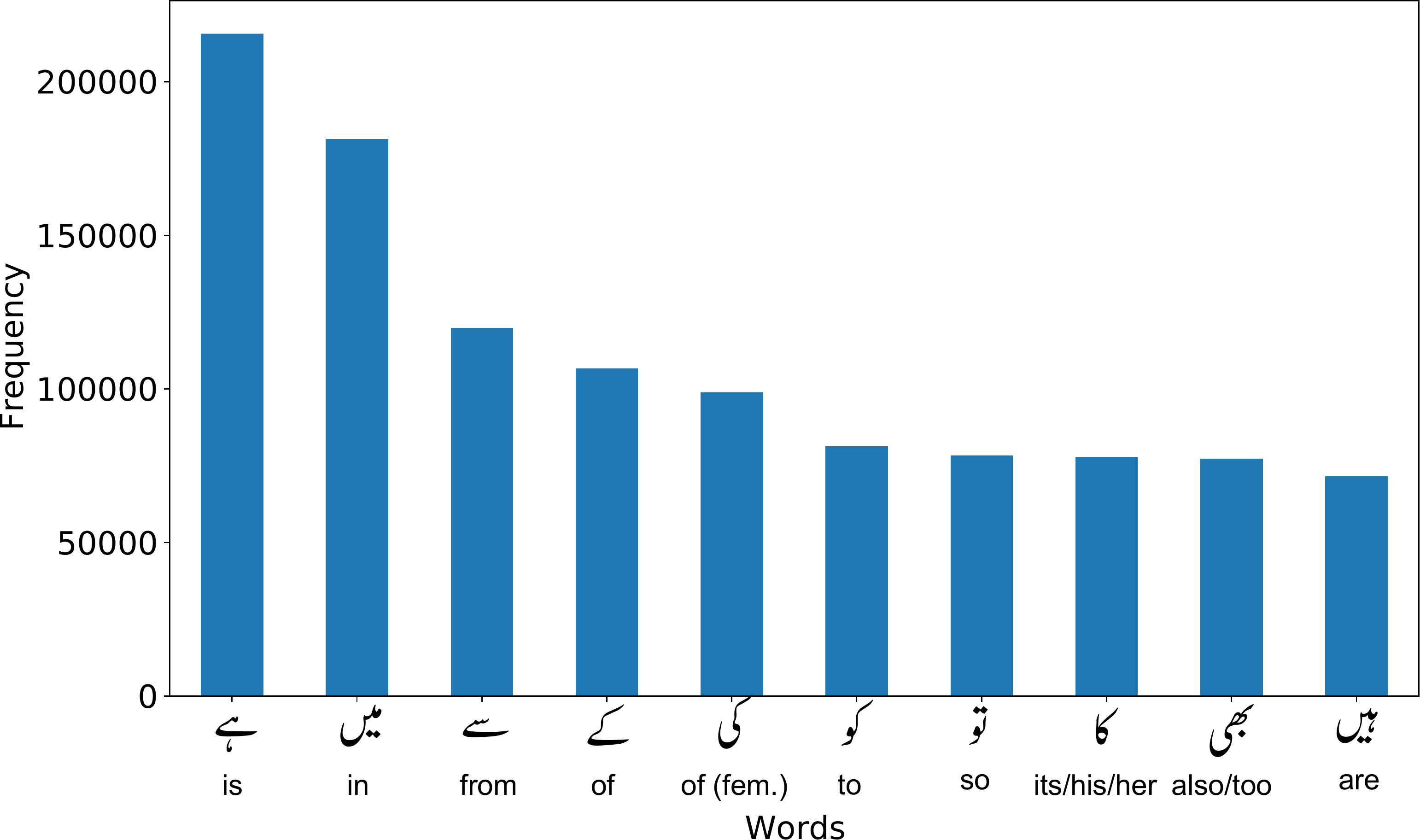}\\
\end{center}
\caption{Frequency of most used words without removing stop-words}
\label{fig:words_WOR}
\end{figure}

\begin{figure}[htbp]
\begin{center}
\includegraphics[scale=0.42]{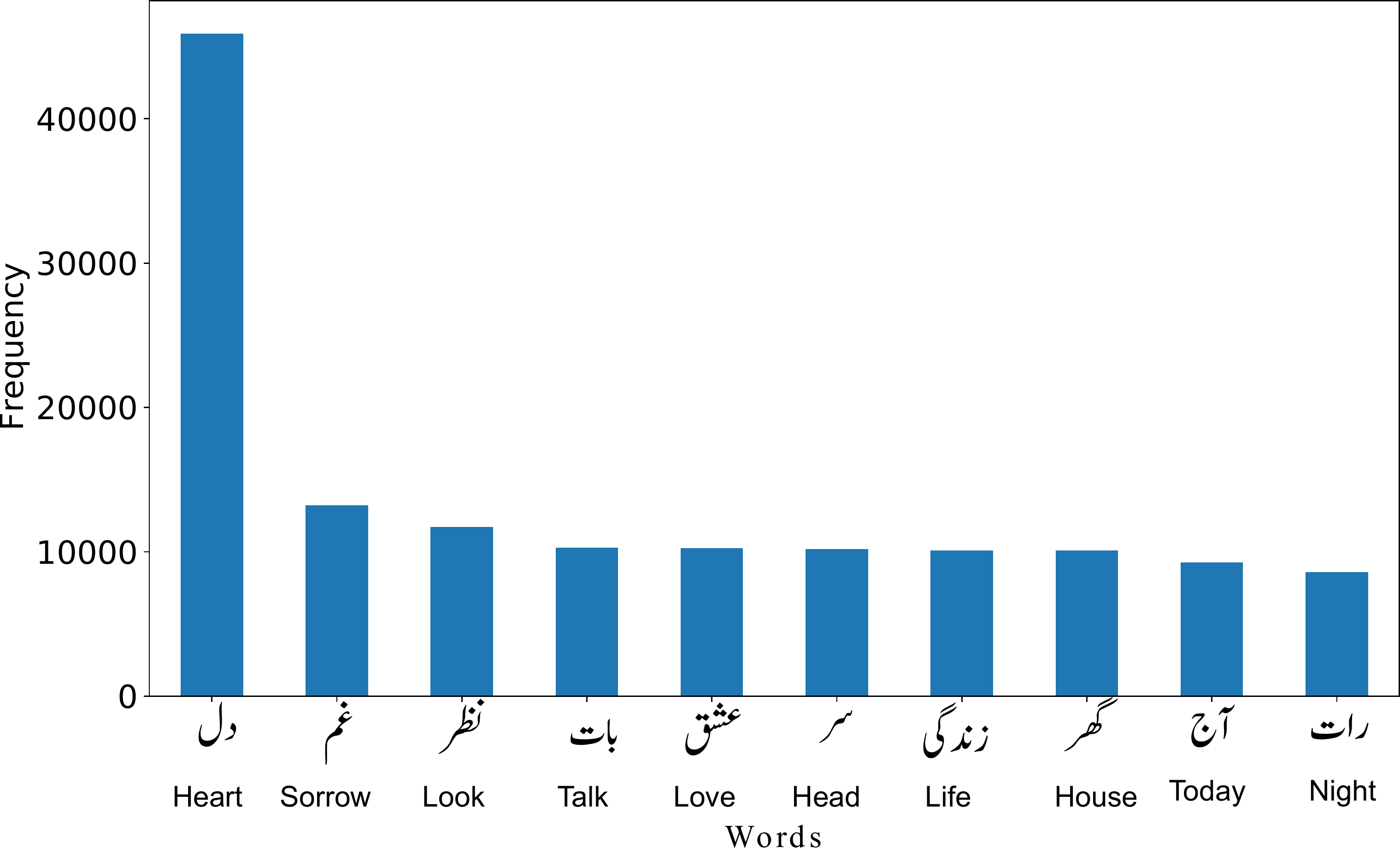}\\
\end{center}
\caption{Frequency of most used words after removing stop-words}
\label{fig:words_WSR}
\end{figure}

Figure \ref{fig:words_WOR} and Figure \ref{fig:words_WSR} depict the most used words; both with and after removing stop-words respectively. It merits mention here that stop words are the words  that do not give useful information unless combined with other words. Without the removal of such stop-words as shown in Figure \ref{fig:words_WOR}, the most predominant words are essentially mundane words consisting of articles, verbs and pre-positions of Urdu language. This was  expected since these words would always be needed in any sentence formulation. On the contrary, a much deeper understanding is communicated by the data presented in Figure \ref{fig:words_WSR}. It can be seen that a clear poetic bias exists in using the word \emph{dil}  (heart) having a word count of nearly 49,000 words. The second most frequently used word is \emph{gham} (grief/sorrow) having a word count of nearly 15000 i.e., almost one third of the word count of \emph{dil}. It also needs due attention here that the word \emph{dil} occupies a central position in Urdu literature. Same is also the case with several other languages. In fact, a vital aspect of poetry is that it allows the heart to lead the mind rather than the reverse (\cite{Butler2009}). Hence, the disproportionateness of the frequency of the most used words is Urdu poetry is both expected and justified.

\begin{figure}[htbp]
\begin{center}
\hbox{\hspace{30ex}\includegraphics[scale=0.42]{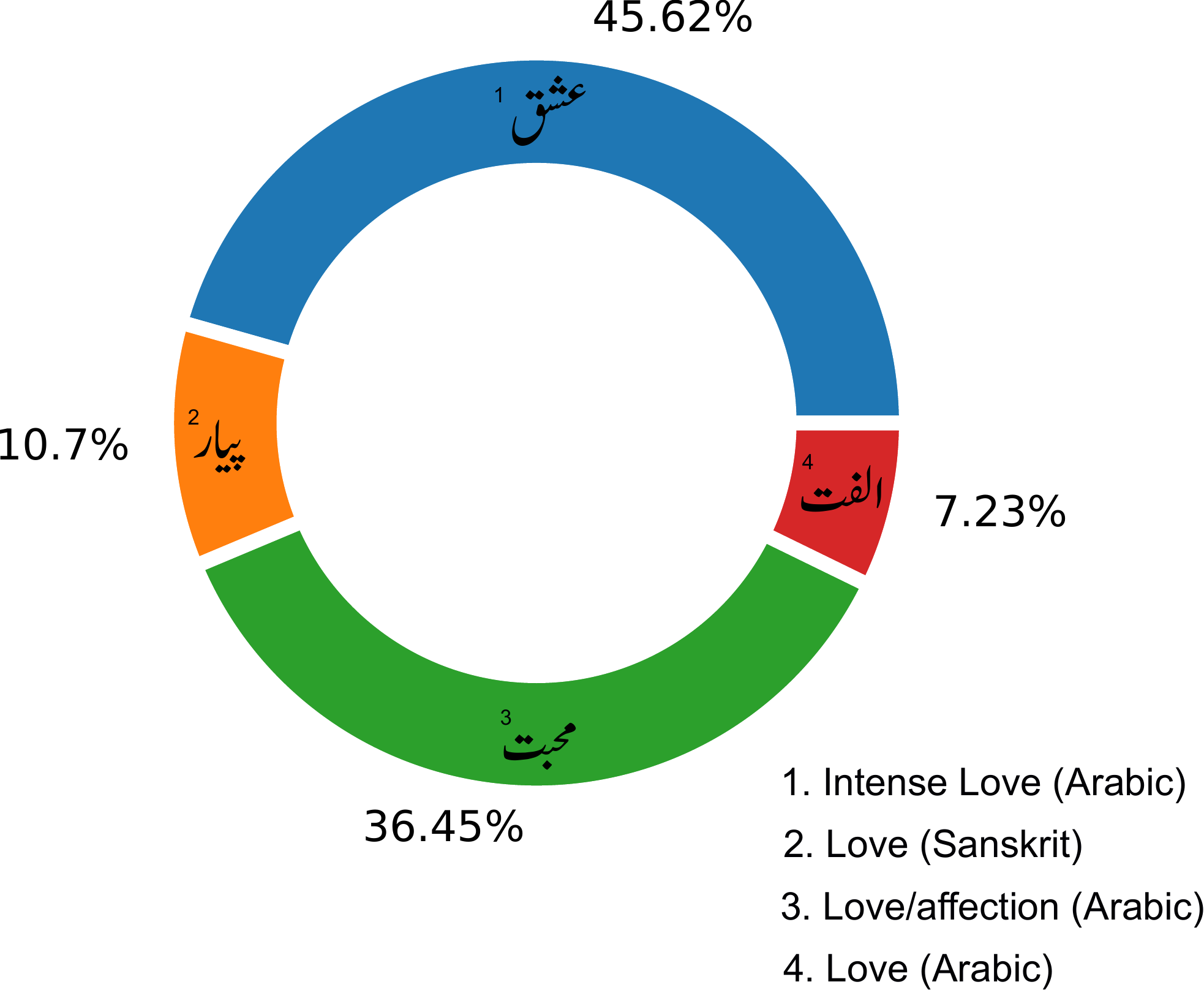}}
\end{center}
\caption{Percentage of different words mentioning with expression of love}
\label{fig:love_words}
\end{figure}

In this study, a categorization of the words frequently used by poets (love, body parts, flowers, birds, and natural habitats, in particular) have also been  performed. In Urdu language, love can be expressed by a number of words like  \emph{‘ishq’} (intense love: \emph{Arabic}),  \emph{‘pyar’} (love: \emph{Sanskrit}), \emph{‘mohabbat’} (love/affection \emph{Arabic}), and \emph{‘ulfat’} (love: \emph{Arabic}). The most commonly used word for expression of love, however, remain \emph{‘ishq’} with a value of 45.62\% among all the words used for expressing love (Figure \ref{fig:love_words}). This is owing to the fact that the word \emph{‘ishq’}  has been both attributed to the love of beloved one as well as the love of God in Urdu poetry. On the other hand, the other words are almost always attributed to the love of ones beloved person only.

Similarly, in the word count for words representing body parts of beloved ones, \emph{‘ankh’} (eye: \emph{Sanskrit}) has been used the most (27.01\%). A rather interesting observation is that a synonym for eye in Urdu \emph{chashm} (eye: \emph{Sanskrit}) has also been used extensively (18.33\%). Hence, there is a clear preference for one word over its synonym. Same is the case with word \emph{lub} (lips: \emph{Persian}) and its synonym \emph{hont} (lips: \emph{Hindi}). There is a preference for the word \emph{lub} (19.58\%) while only a small portion (2.36\%) for \emph{hont}. Other commonly used body parts include \emph{chehra} (face: \emph{Persian}), and \emph{zulf}  (hair lock). All of these words are typically used in poetry to describe the beauty of one's beloved. As well as that, it's also a common practice in poetry to draw parallels between beloved's body parts and other beautiful and tangible things.

\begin{figure}[htbp]
\begin{center}
\end{center}
\hbox{\hspace{27ex}\includegraphics[scale=0.42]{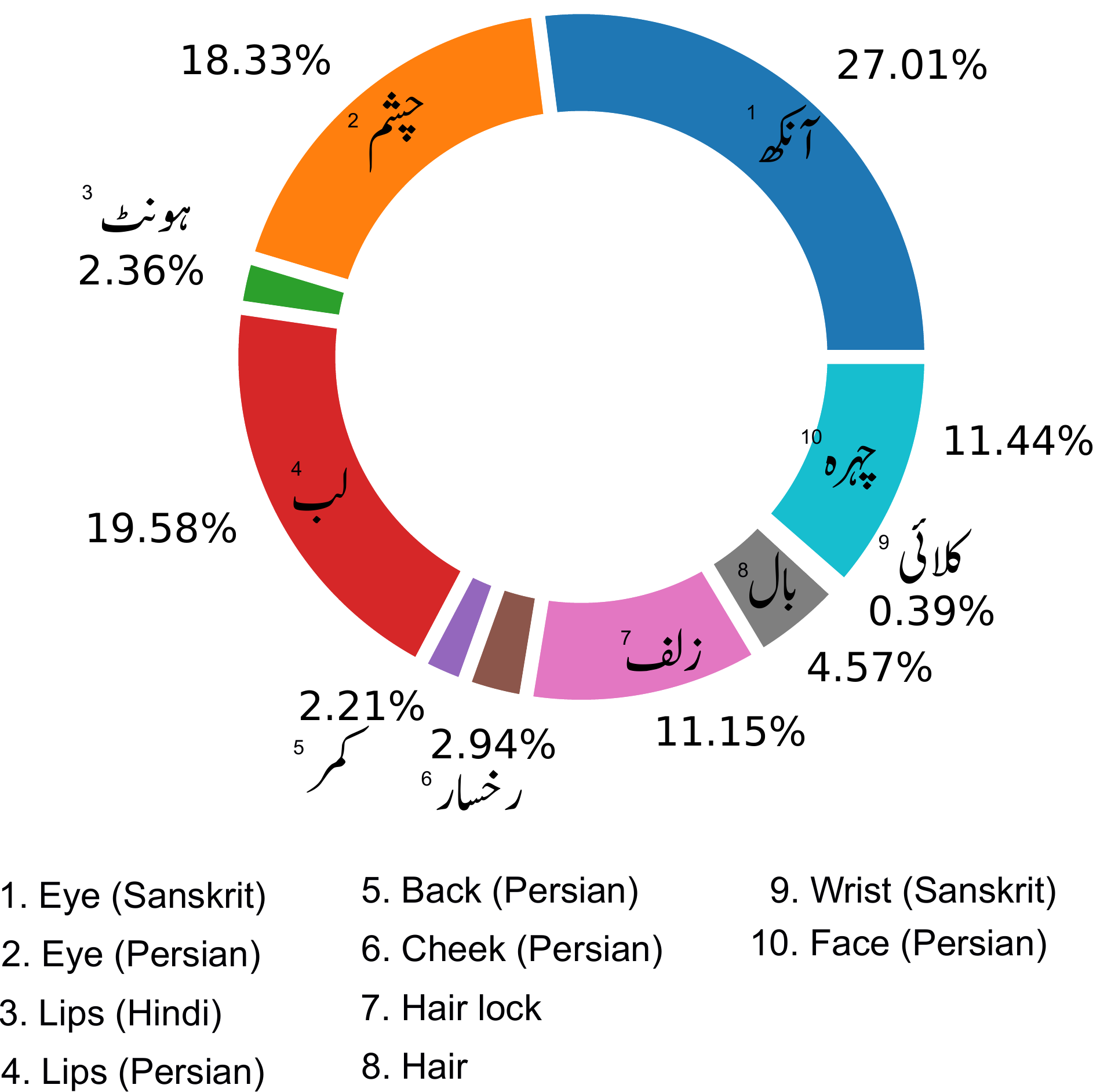}}
\caption{Percentage of different words mentioning body parts of loved one}
\label{fig:body_parts_words}
\end{figure}

\begin{figure}[htbp]
\begin{center}
\hbox{\hspace{33ex}\includegraphics[scale=0.42]{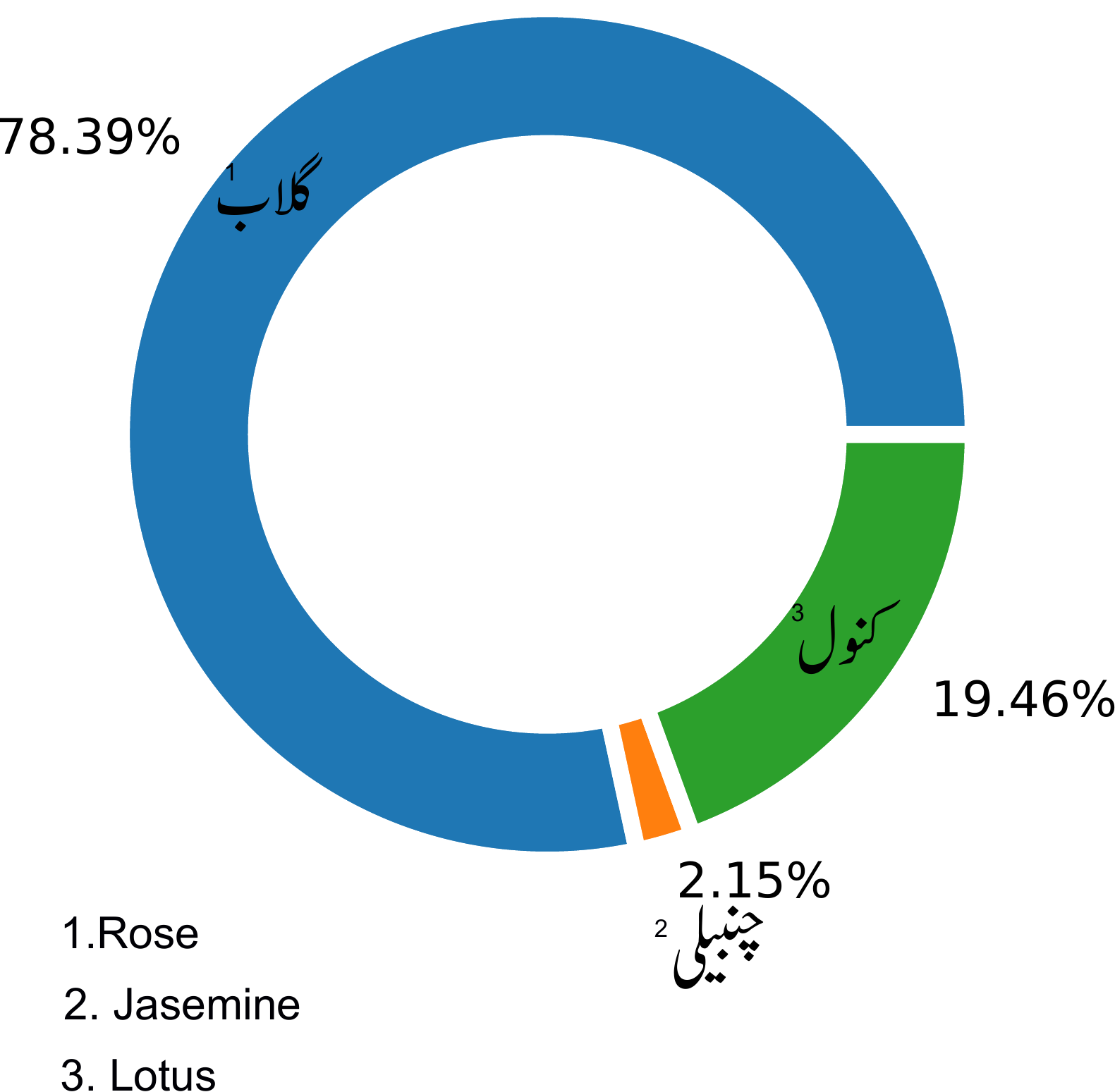}}
\end{center}
\caption{Percentage of  words mentioning names of different flowers}
\label{fig:flower_words}
\end{figure}

In the flowers category, \emph{gulab} (rose) occupies a central position in Urdu poetry. Its beauty, fragrance, and the presence of prickles makes it ideal as a metaphor. Hence, it is no surprise that the word \emph{gulab} constitutes 78.4\% of all the flowers that have been used in the Urdu poetry (Figure \ref{fig:flower_words}). There is an interesting parallel with English poetry here, due to the extensive use of rose in English poetry (\cite{Li2021}). Another stark similarity stems from the fact that \emph{kanwal} (lotus/water lily) comes out as the second most used word (19.46\%), a word that is used the most in Chinese poetry as a symbol of beauty, love and rectitude (\cite{Li2021,Hou}). 

In European literature (more specifically to Greek and Latin literature), the use of nightingale plays an important role as compared to any other bird from poets like Homer to T S Eliot (\cite{Chandler1934}). This is owing to the beautiful look but more importantly all the more beautiful voice of the bird that is considered melodious. Hence, it is no surprise that the word \emph{bulbul} (nightingale) occurs most frequently (52.05\%) in the Urdu poetry data under consideration. This is also testament to the fact the bird is considered a poetic tool not only in the Europe but also in South Asia owing to the reasons already mentioned.

\begin{figure}[htbp]
\begin{center}
\hbox{\hspace{31ex}\includegraphics[scale=0.42]{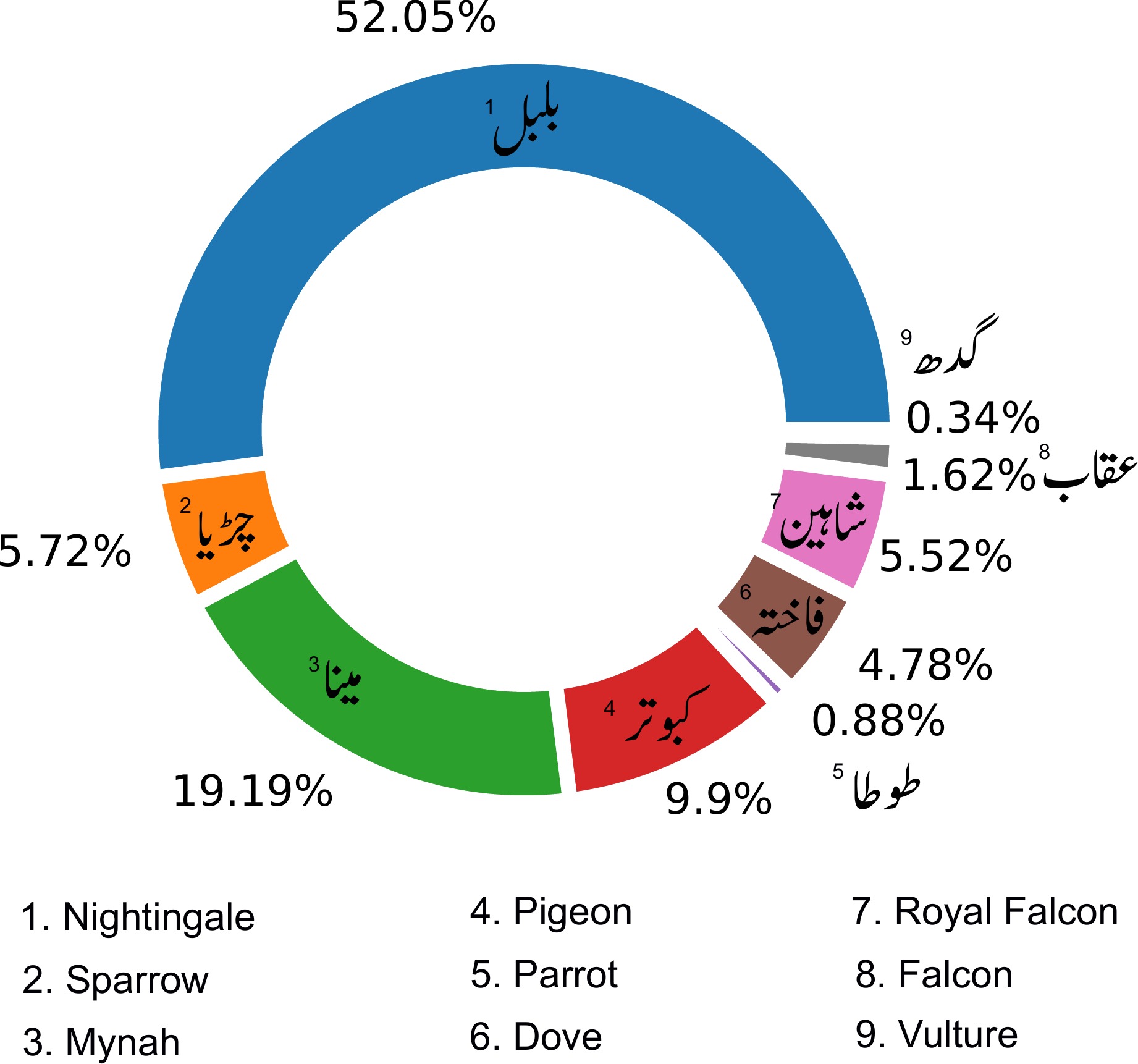}}
\end{center}
\caption{Percentage of words mentioning birds}
\label{fig:Birds}
\end{figure}

\begin{figure}[htbp]
\begin{center}
\hbox{\hspace{28ex}\includegraphics[scale=0.42]{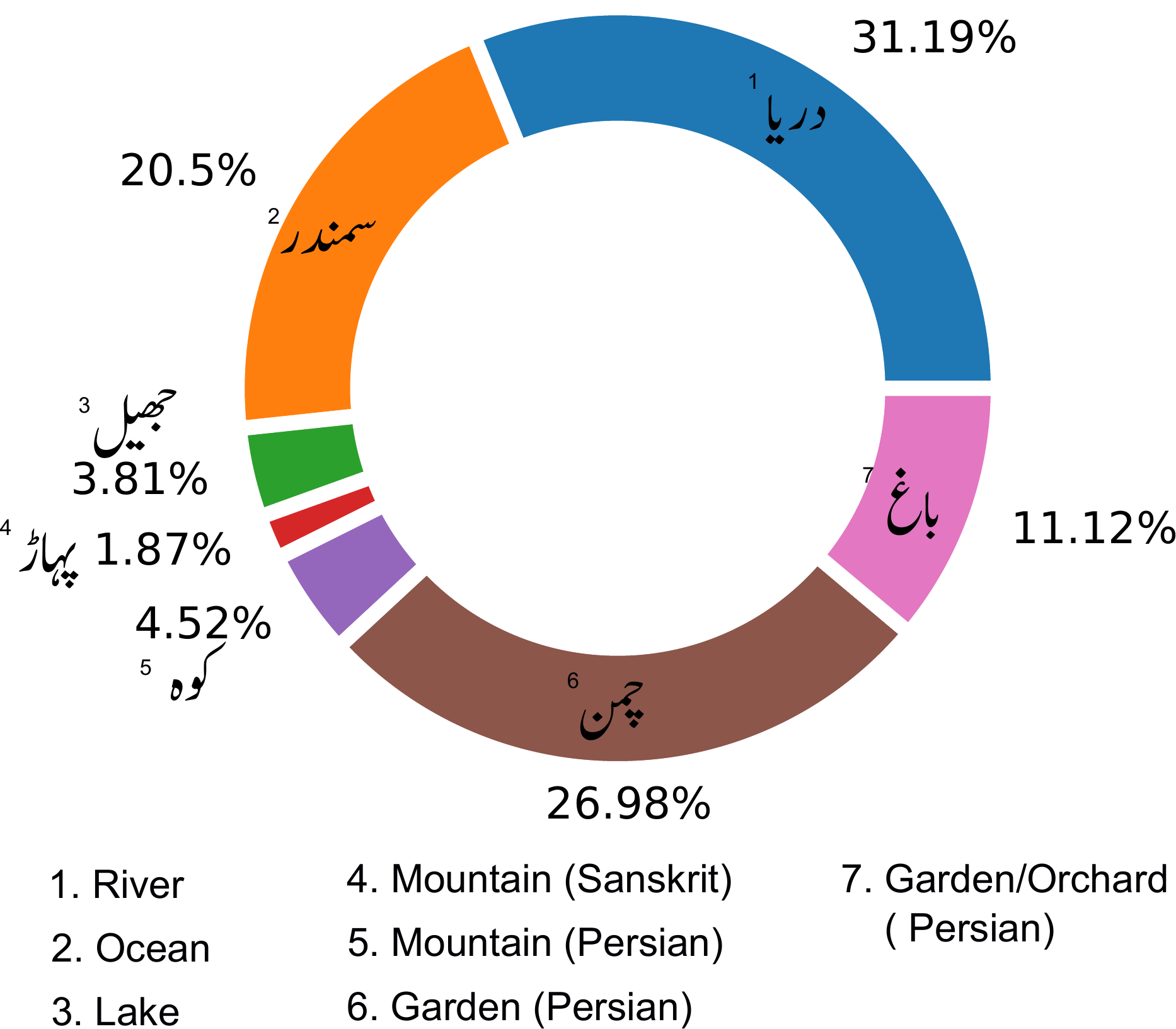}}
\end{center}
\caption{Percentage of  words mentioning different natural habitats}
\label{fig:nature_words}
\end{figure}

Figure \ref{fig:nature_words} shows the percentage plot of words depicting natural habitats. The habitats are also frequently expressed in Urdu poetry. However, there is no bias in the utilization of habitats in the poetry unlike the few biases that were observed in other words. The word \emph{darya} (river) appears most frequently (31.19\%), followed by \emph{chaman} (garden), and \emph{samandar} (ocean).

\begin{figure}[htbp]
\begin{center}
\includegraphics[scale=0.42]{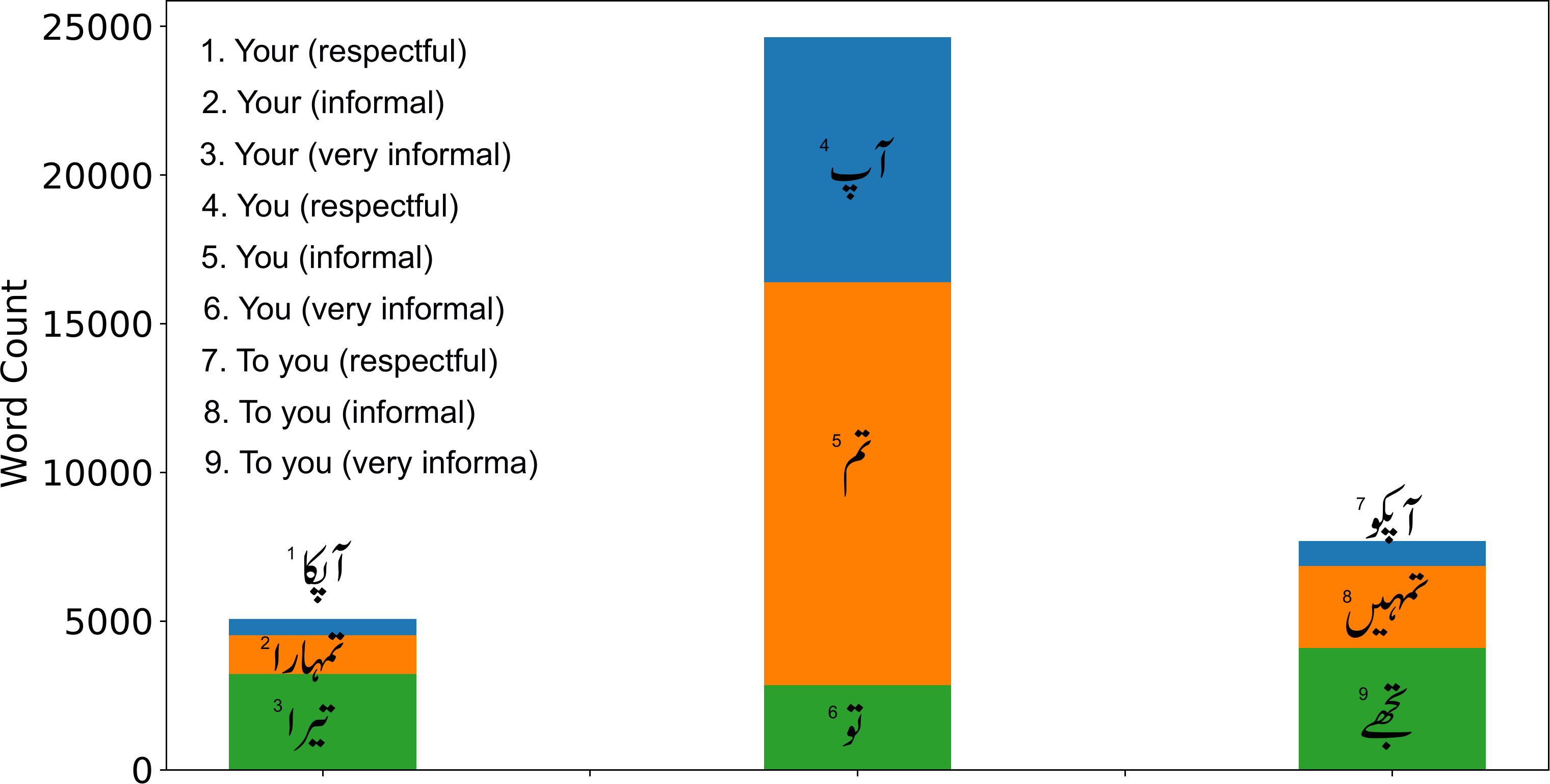}\\
\end{center}
\caption{Count of words of sophistication in addressing the second person}
\label{fig:Second_Person}
\end{figure}

Another important percentage distribution is shown in Figure \ref{fig:Second_Person} where the words used for addressing the second person are plotted against their respective word count. The word ‘you’ in Urdu can be expressed in different ways. For example, \emph{aap’} (you) is considered more respectful, whereas \emph{tum} (you) and \emph{tu} (You) are considered more informal/casual. This is unlike English language where second person is only communicated by the word ‘you’ regardless of the underlying emotion (respect/formal-casual relationship). The word \emph{tum} (you - informal) is used the most as it is informal and most of the poetic formulations are often intended to be between a lover and his beloved one. On the contrary, \emph{tu} (you - very informal) is utilized the least as often \emph{tu} is also considered a bit derogatory. So, it’s less occurrence in the poetry is justified.

\subsection{\emph{Radeef} Words}

\begin{figure}[htbp]
\begin{center}
\includegraphics[scale=0.42]{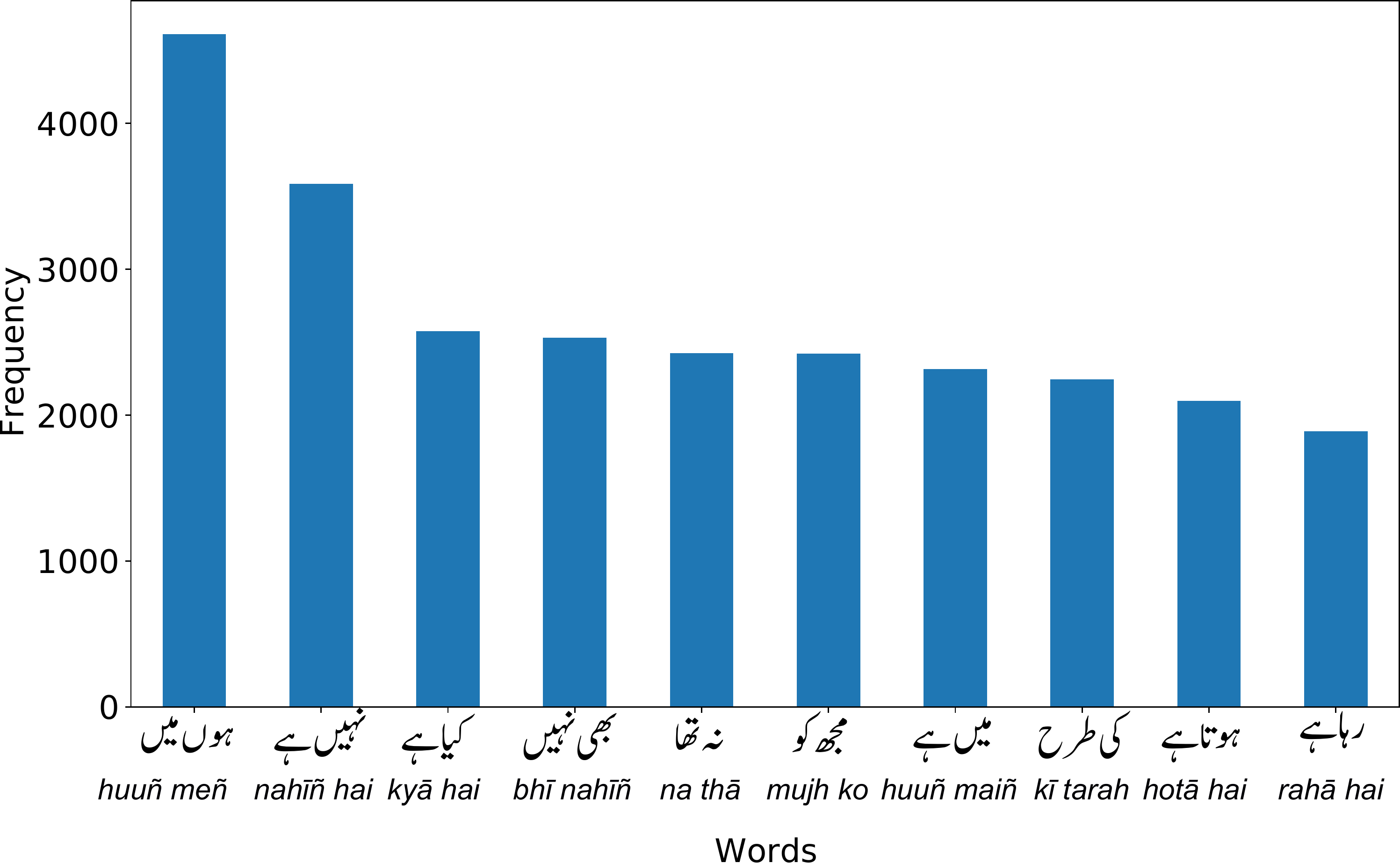}\\
\end{center}
\caption{Frequency of Radeef words in the corpus.}
\label{fig:Qafia}
\end{figure}

\emph{Radeef} is a rule in Urdu poetry where a poet uses same last words  in every second verse of that couplet. There is a dominating application of this rule in rhyming Urdu poetry.  Figure \ref{fig:Qafia} represents the histogram of preferred most by Urdu poets.It is apparent that the dominant rhyming words are those that end with  \emph{huun me} (I am) which shows that Urdu poets are typically verbose about their unique characteristics. This is followed by the word \emph{nahin hai} (is not). It can also be noticed that out of ten most occurring \emph{radeef} words, three  words contain words \emph{nahe, na} (no, not) which brings the expression of negativitiy in the poetry. It should come as no surprise, as the \emph{ghazal} genre is also known for its extreme view of the poet's disapproval by loved ones and society as well.

\subsection{Poetry style}

Multi Dimensional Scaling (MDS) is an AI tool to understand the similarities/difference between data sets based on its inherent features. In MDS, each pair of points in the original high-dimensional space is calculated and then mapped to a lower-dimensional space while preserving the distance between each pair such that similar data sets cluster together. Although this technique is mainly used to find connection between data containing numbers, however more recently this this method has been applied to find similarity between poetic styles of different English poets (\cite{jacobs2018gutenberg}).

\begin{figure}[htbp]
\begin{center}
\includegraphics[scale=0.45]{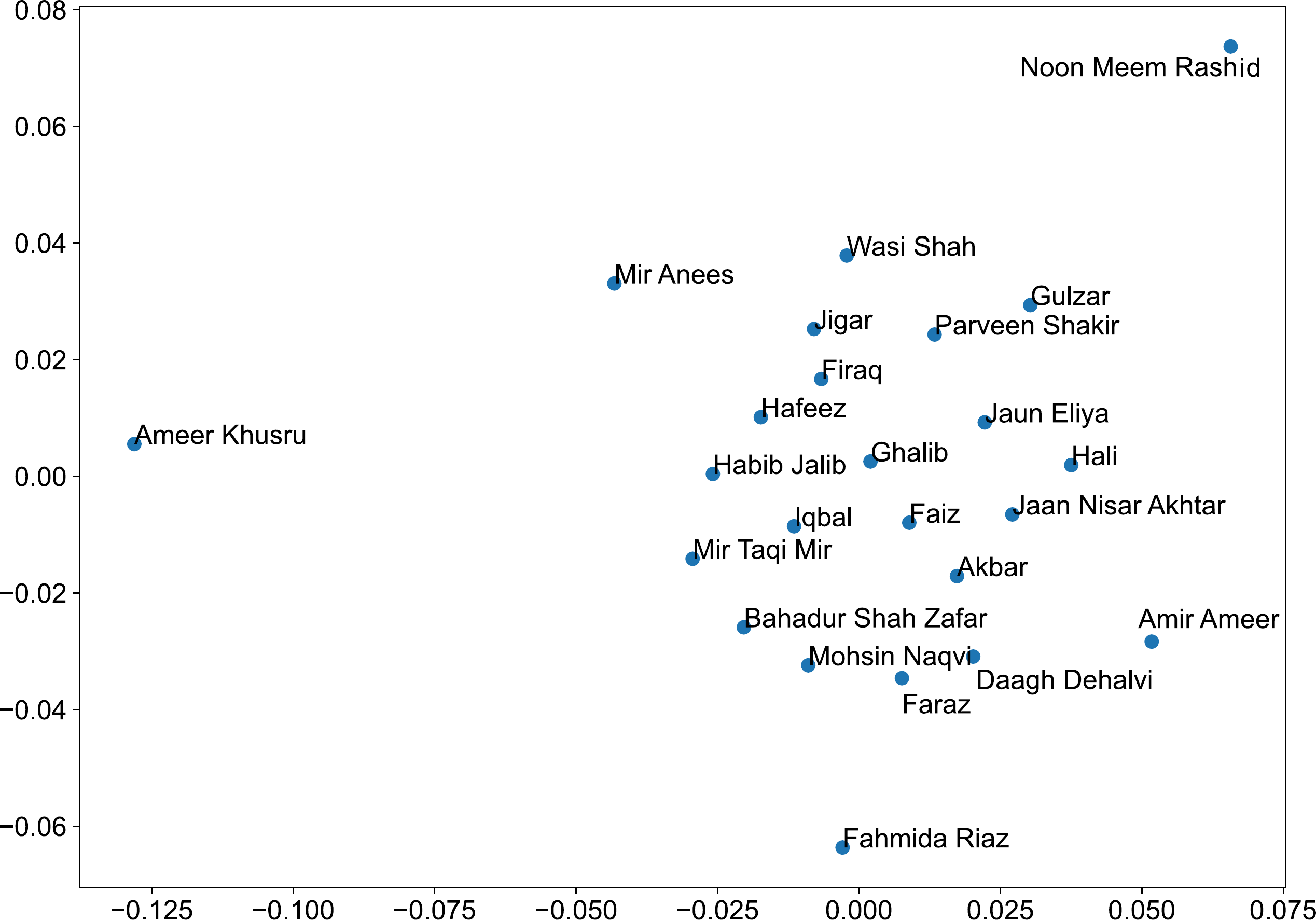}\\
\end{center}
\caption{Multi Dimensional Scaling (MDS) plot of text similarity for selected 24 poets on the basis of latent semantic analysis.}
\label{fig:Poets_MDS}
\end{figure}

Figure \ref{fig:Poets_MDS} shows the MDS plot of 24 prominent Urdu poets . It can be seen that there are a few outliers like Ameer Khusru, N M Rashid, Mir Anees and Fahmida Riaz . Ameer Khusru of Delhi was one of the greatest poets of medieval India. He wrote in both Persian, the courtly language of his time, and Hindavi, the language of the masses. The same Hindavi later developed into two beautiful languages called Urdu and Hindi (\cite{Mysticpoet}). Due to rather archaic word usage by Ameer Khusru which became outdated later, his poetic work becomes a natural outlier. On the other hand, N M Rashid’s work is also an atypical owing solely to his own signature style of poetry. N M Rashid made some outstanding contributions endowing free verse a place of prominence (\cite{Ahmed2008}. Similarly, Mir Anees also becomes an outlier (albeit just marginally) because Mir Anees is known for his works in ‘Marsia-Nigari’ (a form of Elegy in Shia Islam attributed to the Martyrs of Karbala (\cite{Hussain2005}). Due to this peculiar form of poetry, the outlying position of Mir Anees is totally justified. The female poet Fahmida Riaz also stands out because of her unique feminist poems. 

\section{Summary of Key Findings and Way Forward}
In this work, we presented an exploratory analysis of Urdu language. Firstly, the existing literature on Urdu language process using available literature was presented. Later, the scheme for performing EDA of Urdu language poetry was explained. Finally, the results obtained from the EDA were presented and discussed. The results of this study provide an anatomy of existing Urdu poetry and offer insight into its usage of key words imbued with romanticism, analogy, and imagination. Furthermore,as part of our study, the poetic styles of a number of iconic Urdu poets were analyzed using MDS in order to gain a picture of how those poets are similar or divergent.

This study opens up different aspects of Urdu poetry to be analyzed; e.g. to study the usage of different words in chronological order over past few centuries, comparison of \emph{Dahalvi} and \emph{Lakhnavi} Urdu classical poetry which still remains a  debate in literary circles, evolution of poetic work of a poet in different periods of his life etc. Moreover, the findings of this study can be applied to the generation of Urdu poetry using NLP, to the stylization of poetry by computers and to research in neurocognitive poetics.

\bibliographystyle{unsrtnat}
\bibliography{references}  






\end{document}